\pdfoutput=1

\documentclass[11pt]{article}

\usepackage[final]{acl}

\usepackage{times}
\usepackage{latexsym}
\usepackage{amssymb}
\usepackage{pgfplots}
\usepackage{multirow}
\usepackage{scalefnt}
\usepackage{boldline}
\usepackage{colortbl}
\usepackage{makecell}
\usepackage{graphicx}
\usepackage{bm}
\usepackage{amsmath}
\usepackage{bbm}
\usepackage{float}
\usepackage{stfloats}
\usepackage{lipsum}
\usepackage{comment}
\usepackage{booktabs}
\usepackage{subfig}
\usepackage{lipsum}
\usepackage{cleveref}
\usepackage{amsmath}
\usepackage{mathtools}
\usepackage{verbatim}
\usepackage{longtable}
\usepackage{array}
\usepackage{pifont}

\usepackage[flushleft]{threeparttable}

\crefformat{section}{\S#2#1#3} 
\crefformat{subsection}{\S#2#1#3}

\usepackage{booktabs}
\usepackage[T1]{fontenc}

\usepackage[utf8]{inputenc}

\usepackage{microtype}

\usepackage{inconsolata}

\usepackage{tabularx}

\usepackage{xspace}

\usepackage{hyperref}
\usepackage{listings}
\usepackage{color}
\usepackage{xspace}
\usepackage[normalem]{ulem}
\useunder{\uline}{\ul}{}

\definecolor{backcolour}{rgb}{0.95,0.95,0.92}

\definecolor{purple1}{HTML}{7570b3}
\definecolor{green1}{HTML}{1b9e77}
\definecolor{orange}{HTML}{d95f02}
\definecolor{green2}{HTML}{4daf4a}
\definecolor{red}{HTML}{e41a1c}
\definecolor{blue}{HTML}{377eb8}
\definecolor{purple2}{HTML}{984ea3}

\newcommand{\ours}{Generative Prompt Internalization\xspace}
\newcommand{\OURS}{GenPI\xspace}
\newcommand{\datagen}{Self Role-Playing Conversation\xspace}
\newcommand{\stageone}{Prompt Generation loss\xspace}
\newcommand{\STAGEONE}{PG\xspace}
\newcommand{\stagetwo}{SFT loss\xspace}
\newcommand{\STAGETWO}{SFT\xspace}

\newcommand{\cmark}{\ding{51}}
\newcommand{\xmark}{\ding{55}}

\lstdefinestyle{task_prompt}{
    basicstyle=\rmfamily\tiny,
    breakatwhitespace=true,
    breaklines=true,
    breakautoindent=false,
    breakindent=0pt
}

\lstdefinestyle{data-example}{
    basicstyle=\rmfamily\scriptsize,
    breakatwhitespace=true,
    breaklines=true,
    breakautoindent=false,
    breakindent=0pt
}

\lstdefinestyle{component-prompt}{
    basicstyle=\rmfamily\scriptsize,
    breakatwhitespace=true,
    breaklines=true,
    breakautoindent=false,
    breakindent=0pt
}

\title{\ours}

\author{
    Haebin Shin$^{1,2}${\thanks{\, \* Work done during internship at Microsoft Research}}
        \quad
    Lei Ji$^2$
        \quad
    Yeyun Gong$^2$ 
        \quad
    Sungdong Kim$^{1}$
        \hspace{6pt}
    Eunbi Choi
        \hspace{6pt}
    Minjoon Seo$^1$ 
    \\
    \\
    $^1$KAIST AI \quad $^2$Microsoft Research
    \\[3pt]
    \texttt{\{haebin.shin, minjoon\}@kaist.ac.kr}
}

\begin{document}
\maketitle

\begin{abstract}




Prompts used in recent large language model based applications are often fixed and lengthy, leading to significant computational overhead. To address this challenge, we propose \ours (\OURS), a lightweight method that employs a joint training approach. \OURS not only replicates the behavior of models with prompt inputs but also generates the content of the prompt along with reasons for why the model's behavior should change accordingly. We demonstrate that our approach effectively internalizes complex prompts across various agent-based application scenarios. 
For effective training without interactions with the dedicated environments, we introduce a data synthesis technique that autonomously collects conversational datasets by swapping the roles of the agent and environment. This method is especially useful in scenarios where only a predefined prompt is available without a corresponding training dataset. By internalizing complex prompts, \ours enables high-performance and efficient inference without the need for explicit prompts.\footnote{Code are available at \href{https://github.com/kaistai/GenPI}{https://github.com/kaistai/GenPI}}.



%

\end{abstract}

\section{Introduction}
\label{sec:intro}


In the real world, for inference in large language model (LLM) based applications (e.g. ChatGPT), fixed and complex prompts are often used repeatedly.
Although advanced prompts can improve performance, their dependence on lengthy prompts raises concerns regarding computational efficiency for service providers and cost-effectiveness for users. Longer prompts can considerably increase computational overhead, especially during multi-turn inference, making it a potential bottleneck in practical applications. Therefore, finding effective strategies to optimize these prompts while maintaining performance is crucial for the practical deployment of LLM-based applications. 

To address this issue, existing methods can be categorized into two approaches.
One practical approach only for efficient prompt computation is compressing prompts into external embeddings~\citep{chevalier-etal-2023-adapting, mu2024learningcompresspromptsgist, ge2024incontext} or compressed text ~\citep{jiang-etal-2023-llmlingua, li-etal-2023-compressing_selective, jiang-etal-2024-longllmlingua, pan-etal-2024-llmlingua}.
However, these methods still require additional tokens, limiting their ability to fully internalize prompts and effectively modify the model’s behavior as intended. Alternatively, fine-tuning~\citep{kim-rush-2016-sequence, zou2024promptinternsavinginferencecosts} or distillation~\citep{askell2021generallanguageassistantlaboratory, snell2022learningdistillingcontext, choi-etal-2023-fixed, li2024mendmetademonstrationdistillation}, have been explored as canonical approaches for internalizing prompts. 
These methods adjust the language model’s behavior to follow the prompt’s intentions without requiring a prompt during inference.

However, these internalization methods are limited in that the model cannot reference the prompt's content during training. Instead, they rely on indirect training based on the model's output~\citep{kim-rush-2016-sequence} or distribution~\citep{askell2021generallanguageassistantlaboratory, snell2022learningdistillingcontext, choi-etal-2023-fixed, li2024mendmetademonstrationdistillation} of the original model when the prompt is provided.
This leads to significant performance degradation in information-rich tasks or specific requirements, such as agent's action requirements or schema grounding~\citep{choi-etal-2023-fixed}.

To overcome the limitations, we propose \ours (\OURS), a method that is trained to generate the target prompt, rather than merely using it as input. \OURS employs joint loss training, combining two approaches: 1) mimicking the teacher's output to guide behavior, similar to distillation approaches~\citep{askell2021generallanguageassistantlaboratory, snell2022learningdistillingcontext, choi-etal-2023-fixed, li2024mendmetademonstrationdistillation}, and 2) generating the content of the prompt while inferring why the output should change based on that prompt. \OURS employs a lightweight internalization process for each prompt, requiring only 0.5\% additional parameters on 1,000 samples.
Focusing on scenario where only prompts are available without a corresponding training dataset for internalization~\citep{liu2023agentbenchevaluatingllmsagents, choi-etal-2023-fixed}, we also introduce a data synthesis technique to generate a multi-turn pseudo conversational dataset.
By simply swapping the roles of user and assistant within the given context, the model simulates both sides of conversation, enabling it to autonomously collect conversational datasets.

We evaluate our method in agent application scenarios with lengthy prompts using AgentBench~\citep{liu2023agentbenchevaluatingllmsagents}. \OURS maintains strong performance even without the need for prompt input, while outperforming other distillation or compression-based baselines. 
\OURS achieves 100\% performance retention on OS interaction agent task, while maintaining at least 82\% performance on web-based agent tasks with over 1,000 tokens.
Furthermore, \OURS demonstrates 39\% efficiency improvement in handling environments with lengthy prompts, outperforming other compression-based methods.

\vspace{-0.1em}
\section{Related Works}
\label{sec:related_work}
\vspace{-0.3em}
\paragraph{Prompt Internalization} \hspace{-0.7em} is a methodology designed to embed prompt information within a language model, enabling the model to perform various tasks without requiring explicit prompt input. For instance,~\citet{askell2021generallanguageassistantlaboratory} internalize persona-related prompts to facilitate helpful, honest, and harmless alignment, while~\citet{snell2022learningdistillingcontext} incorporates both prompts and scratch pads into the internalization process to enhance performance on more complex tasks. \citet{li2024mendmetademonstrationdistillation} applies prompt internalization to internalize demonstrations for in-context learning scenarios. While these methods focus on internalizing broad and general coarse-grained information,~\citet{choi-etal-2023-fixed, zou2024promptinternsavinginferencecosts} introduce a more fine-grained approach, targeting more specific and predetermined prompts. ~\citet{choi-etal-2023-fixed} focuses on short chat histories or task-specific instructions, and~\citet{zou2024promptinternsavinginferencecosts} relies on a tailored training dataset to retrieve similar examples based on the given user input.
Following these previous works, our goal is to enable the model to internalize the specific predetermined prompts. However, existing methods still limit the model's ability to directly learn the content of the prompt, as they rely on training a student model using outputs from a teacher model. We address this limitation by generating prompts that allow the model to learn their content directly, enabling it to handle more realistic and information-rich prompts.


\vspace{-0.3em}
\paragraph{Prompt Compression} \hspace{-0.7em} is one of the practical approach to reducing the computational overhead caused by lengthy prompts.
In case of the API-based large language models (LLMs) services,  systematic caching solutions allow frequently used prompts to be stored between API calls.\footnote{https://www.anthropic.com/news/prompt-caching}\footnote{https://ai.google.dev/gemini-api/docs/caching}
For users of API-based LLMs, text-based prompt compression methods are proposed, where key segments of long prompts are selected on a token-by-token basis to generate a compressed version of the original prompt~\citep{jiang-etal-2023-llmlingua, li-etal-2023-compressing_selective, jiang-etal-2024-longllmlingua, pan-etal-2024-llmlingua}. Although this approach reduces prompt length, it often still results in relatively long token sequences, as essential tokens must be retained.
On the other hand, embedding-based prompt compression methods generate cached token embeddings for prompts, which can be utilized as a more efficient representation in LLMs~\citep{chevalier-etal-2023-adapting, mu2024learningcompresspromptsgist, ge2024incontext}. These methods offer the advantage of using fewer token embeddings compared to text-based compression but often require modifications to the model architecture, making it challenging to leverage the compressed vectors across different models.

\section{Problem Definition}
\vspace{-0.2em}
\label{sec:definition}


Following \citet{choi-etal-2023-fixed}, we assume a scenario where an application-specific prompt $p$ is predetermined. Our goal is to guide the model to behave as if the prompt is given, even in its absence. Similar to the previous distillation approaches~\citep{askell2021generallanguageassistantlaboratory, snell2022learningdistillingcontext, choi-etal-2023-fixed, li2024mendmetademonstrationdistillation}, we define the teacher model $T$ and the student model $S$ based on whether the prompt $p$ is provided to the same model $\theta$.

The teacher model $T$ is defined as a function that takes the prompt $p$ and the input $x_i$ at turn $i$, generating the teacher's output:
\begin{math}
    y_{i}^{T} = T(x_i,p) = f_{T}(x_i,p;\theta).
\end{math}
The student model $S$ is a function that takes the input $x_i$ at turn $i$, generating the student's output:
\begin{math}
    y_{i}^{S} = S(x_i) = f_{S}(x_i;\theta).
\end{math}
We approximate the student model's behavior to match that of the teacher model, conditioned on the prompt $p$ over multiple turns, as shown in Equation~\ref{eq:problem_definition},
\begin{equation}
    P(y_i^S|x_i) \approx P(y_i^T|x_i,p).
    \label{eq:problem_definition}
\end{equation}


\section{\ours}
\label{sec:method}
\begin{figure*}[t!]
\centering
\includegraphics[width=1.0\textwidth]{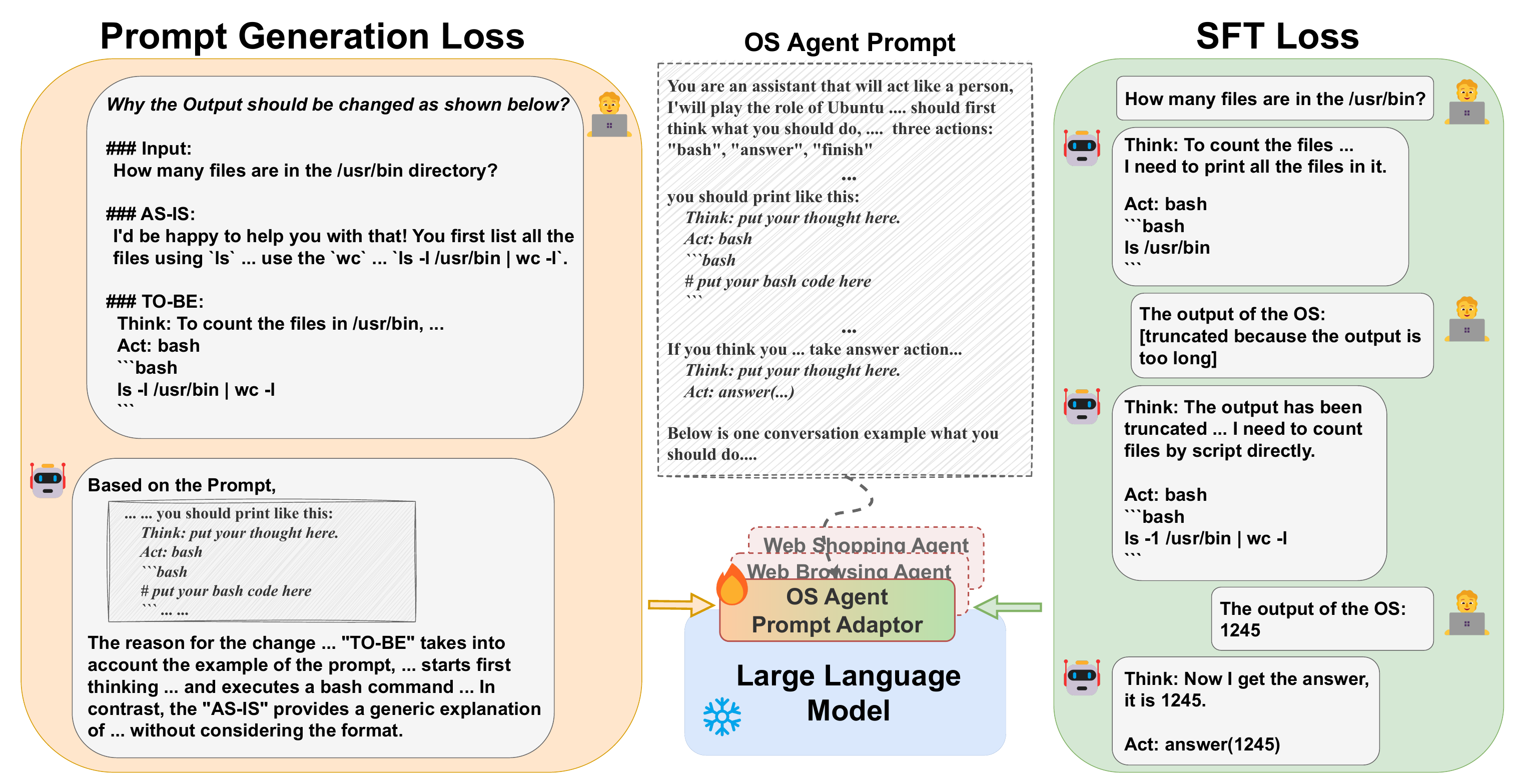}
\caption{Overview of \ours. \stagetwo learns the teacher model’s behavior based on the user input. \stageone internalizes the prompt by generating both the content of the prompt and the reason for why the model’s behavior should be modified. This process is guided by comparing the student model’s output (``AS-IS'') with the teacher model’s output (``TO-BE'').
\stagetwo and \stageone are combined into a joint loss to train the prompt-specific adaptor.}
\label{fig:main}
\end{figure*}


In this section, we demonstrate our novel method, \ours (\OURS), which internalizes the given prompt by generating its own contents. Basically, \OURS involves training the student model to mimic the behavior of the teacher model given the prompt $p$. 
We finetune the student model on the multi-turn outputs generated by the teacher model ($y^T_i, i \in \{0,1,...,N\}$), inspired by sequence-level knowledge distillation approaches~\citep{kim-rush-2016-sequence, touvron2023llama2openfoundation} which follow the hard label distribution from the teacher model. 
\begin{equation}
    \mathcal{L}_{\STAGETWO} = - \sum_{i=0}^{N} \log P(y_i^T|x_{<i},y_{<i}^T,x_i)
    \label{eq:stage2}
\end{equation}

However, the model is still unable to learn the content of the prompt directly; it only learns indirectly through the teacher's output. We introduce an additional loss function, \stageone (\STAGEONE), where the loss is calculated directly on the prompt. This loss involves training the student model to understand why its behavior should align with the teacher's behavior based on the prompt content. 
As illustrated in Figure~\ref{fig:main}, for a given input $x$, the model is trained to generate a prompt $p$ along with a reason $r$ for why the output should be changed. In this process, the student's output $y_i^S$ is considered the initial state (``AS-IS''), while the teacher's output $y_i^T$ represents the desired state (``TO-BE''). This process is formalized in Equation ~\ref{eq:stage1}:
\begin{equation}
    \mathcal{L}_{\STAGEONE} = - \log P(p,r|y_0^S,y_0^T,x).
    \label{eq:stage1}
\end{equation}

We utilize a hyperparameter $\lambda$ to combine the losses into a joint loss function, resulting in the final joint loss function:
\begin{equation}
    \mathcal{L} = (1-\lambda)~\mathcal{L}_{\STAGEONE} + \lambda~\mathcal{L}_{\STAGETWO}
\label{eq:joint_loss}
\end{equation}
    
Our goal is to enable lightweight training and inference for each prompt, while also adapting to changes in the prompt effectively. To achieve this, we employ QLoRA~\citep{dettmers2023qlora} to learn prompt-specific adaptors, allowing us to tailor our approach to each individual prompt.

\section{Components for Prompt Internalization}
\label{sec:component}
Following \citet{choi-etal-2023-fixed}, we assume a realistic scenario where the predetermined prompt has never been encountered before and is not included in any training dataset for prompt internalization. 
Consequently, we first generate the components as a pseudo training dataset from the prompt $p$.

\paragraph{Pseudo User Input.}
Similar to typical query generators~\citep{lewis-etal-2021-paq, choi-etal-2023-fixed,oh-etal-2024-ktrl}, we simply utilize a large language model to generate pseudo user input $x$ by prompting to generate a probably questionable user input from the given prompt $p$. In this paper, we generate only 1,000 pseudo user inputs. Implementation details and examples are provided in~\Cref{appendix:components_implementation_details} and~\Cref{appendix:components_examples}.

\begin{figure}[t!]
\centering
\includegraphics[width=0.85\columnwidth]{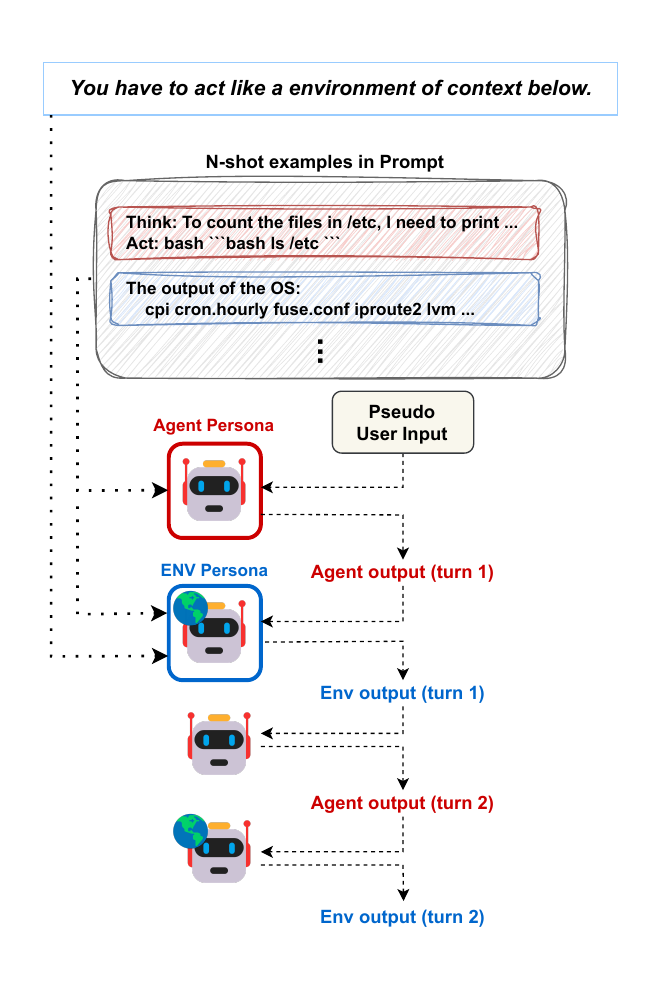}
\caption{Self Role-Playing conversation. Collecting pseudo conversational output by switching the role in the prompt.}
\label{fig:self_role-playing}
\end{figure}

\paragraph{Pseudo Conversational Outputs.}
In many task-specific applications, such as agent applications, the model interacts with the real environment across multiple turns~\citep{liu2023agentbenchevaluatingllmsagents, zeng2023agenttuning, chen2024agentflan}.
This means that collecting multi-turn conversation outputs requires interacting with the actual environment at every turn, which is a non-trivial problem. As our primary objective is to gather the teacher model's behavioral patterns, rather than to optimize the task performance, we introduce a simple method: \textit{\datagen}. This method simply involves reversing the role between agent and environment.
As illustrated in Figure~\ref{fig:self_role-playing}, we provide the model $\theta$ with a role-reversed context to simulate the task environment. By swapping the agent and environment roles in the prompt, a single model $\theta$ can effectively embody both environment and agent personas, enabling the collection of self-conversational outputs $y$. We report the examples and additional quality evaluation in Appendix~\ref{appendix:pseudo_conversational_output_quality}. Implementation details and examples are provided in~\Cref{appendix:components_implementation_details} and~\Cref{appendix:components_examples}.

\paragraph{Reason.}
To collect the supervision for the reason $r$, we utilize a large language model by prompting to generate from the prompt $p$, user input $x$, and outputs $y^S$, $y^T$. 
As illustrated in~\Cref{fig:main}, the reason explains why the student's output (``AS-IS'') should be changed to teacher's output (``TO-BE'').
Implementation details and examples are provided in~\Cref{appendix:components_implementation_details} and~\Cref{appendix:components_examples}.

\section{Experiments Setup}
\label{sec:exp_setup}
\subsection{Dataset}
We evaluate our method on the three agent benchmark tasks: OS-Interaction~\citep{liu2023agentbenchevaluatingllmsagents}, Web Browsing~\citep{deng2023mind2web}, Web Shopping~\citep{yao2022webshop}. As mentioned in Section~\ref{sec:component}, we assume our scenario where an application-specific prompt $p$ is predetermined before the model is deployed. For that, we utilize the AgentBench~\citep{liu2023agentbenchevaluatingllmsagents} settings, which organize a prompt with multi-turn evaluation samples for each task. 
Each task's prompt consists of general task description, agent's action space description, and shot-examples for agent behavior. Detailed information for each task's prompt is described in Appendix~\ref{appendix:dataset_details}.

\paragraph{OS Interaction~\citep{liu2023agentbenchevaluatingllmsagents}.} 
This task involves interacting with an Ubuntu Docker container using bash commands and committing answers, with a 474 token\footnote{All tokens are calculated by the LLaMA tokenizer~\citep{dubey2024llama3herdmodels}.} prompt and 144 evaluation samples. The agent’s performance is measured by the Success Rate.

\paragraph{Web Browsing~\citep{deng2023mind2web}.} 
This task formulates element selection as a multi-choice QA problem. We follow the AgentBench~\citep{liu2023agentbenchevaluatingllmsagents} setting, where the agent operates within an HTML action space (e.g., click, type, select options) using a 1,424 token prompt and 100 evaluation samples. The agent is evaluated using a Success Rate metric, which is based on two criteria: the correctness of the selected element and whether the predicted operation matches the ground truth value for each step.

\paragraph{Web Shopping~\citep{yao2022webshop}.} In this task, the agent navigates scraped \textit{amazon.com} pages to identify the correct product using click or search actions. The task consists of a 1,285 token prompt and 200 evaluation samples.
The task is completed when the agent clicks ``buy now'' or exceeds the turn limit, and performance is evaluated using a reward function that assesses the similarity between the expected and actual product attributes as following the \citet{liu2023agentbenchevaluatingllmsagents}. Please refer to Appendix~\ref{appendix:dataset_details} for more details on the metric.

\subsection{Baselines}
We explore various baselines to internalize the prompt, ranging from distillation approaches~\citep{snell2022learningdistillingcontext, choi-etal-2023-fixed} to compression approaches~\citep{pan-etal-2024-llmlingua, ge2024incontext}.

\paragraph{Full Fine-tuning.}
Since previous typical prompt internalization methods are based on much smaller model~\citep{choi-etal-2023-fixed} or are not publicly available ~\citep{snell2022learningdistillingcontext}, we adopt recent knowledge distillation approaches~\cite{lin2020autoregressiveknowledgedistillationimitation, gu2024minillm, ko2024distillm} for autoregressive large language models in our prompt internalization setup. (1) The basic distillation approach~\citep{Hinton2015DistillingTK} employs the Kullback-Leibler Divergence loss (KLD) between the logit distributions of the student and teacher model. (2) Sequence-level Knowledge Distillation (SeqKD)~\citep{kim-rush-2016-sequence} enforces the student model to generate the teacher model’s outputs on a fixed dataset. (3) As a strong baseline, we also employ a joint loss (SeqKD+KLD), inspired by recent knowledge distillation works~\citep{song2020lightpafftwostagedistillationframework, gu2024minillm}, which incorporates the language modeling loss during distillation. This approach can be interpreted as a hybrid distillation loss that combines the benefits of both soft labels and hard labels from the teacher model.

\paragraph{Prompt Prepending.}
One straightforward approach to consider is prepending the prompt during finetuning. If the prompt is consistently prepended during training, we expect the model to indirectly contextualize the prompt. However, since the model needs to predict without the prompt during inference, there is a potential mismatch between the training and inference distributions. To address this concern, we compare two baseline approaches: (1) always prepending the prompt during training (100\% probability), and (2) stochastically prepending the prompt during training (50\% probability). This baseline utilizes QLoRA~\citep{dettmers2023qlora} with the same settings as our method. This approach can be viewed as an extension of the method that relies exclusively on \stagetwo for training.

\paragraph{Text-based Prompt Compression.}
LLMLingua-2 \citep{pan-etal-2024-llmlingua} is a prompt-agnostic method for generating compressed texts. LLMLingua-2 explicitly compresses tokens using a smaller model, such as XLM-RoBERTa-large~\citep{conneau2020xlm-roberta}. By performing with the compressed prompt, this baseline is expected to achieve efficient inference while maintaining comparable performance.

\paragraph{Embedding-based Prompt Compression.}
We utilize ICAE~\citep{ge2024incontext} to compress the prompt into cached prompt embeddings. Following \citet{ge2024incontext}, we compress the prompt into 128 tokens and prepend compressed tokens to the user input at each inference time. 
Since the official checkpoint of ICAE exhibits significantly low performance on AgentBench~\citep{liu2023agentbenchevaluatingllmsagents}, we finetune the baseline specifically for AgentBench~\citep{liu2023agentbenchevaluatingllmsagents}. Additional details regarding this baseline are described in Appendix~\ref{appendix:embedding_based_prompt_compression_details}.

\paragraph{Upper Bound.}
We utilize the teacher model as an upper bound that inputs the full prompt, consistent with previous studies~\citep{snell2022learningdistillingcontext, choi-etal-2023-fixed}. In our preliminary experiments (see \Cref{appendix:upper_bound_details}), we observe that the fine-tuned model performed worse than the original model in AgentBench~\citep{liu2023agentbenchevaluatingllmsagents}. Consequently, we consider the original model with the full prompt as the teacher model following the setup of~\citet{liu2023agentbenchevaluatingllmsagents}. Since ICAE~\citep{ge2024incontext} is based on the Mistral-7B~\cite{jiang2023mistral7b} model, we establish a separate upper bound specifically for ICAE~\citep{ge2024incontext} to ensure a fair comparison.
For a detailed explanation of the upper bound, please refer to the~\Cref{appendix:upper_bound_details}.

\subsection{Implementation Details}
To internalize the agent-based prompt into the language model, we utilize the LLaMA-3-8B-Instruct~\citep{dubey2024llama3herdmodels} as the target model. Following our scenario in Section~\ref{sec:component}, we fine-tune the model using the 1,000 pairs of pseudo dataset as our train dataset, and this is applied equally to all baselines. 
We utilize QLoRA~\citep{dettmers2023qlora} with rank $r=16$ only requires 0.5\% of parameters.
Additional training details are described in~\Cref{appendix:hyperparameter_details}.

\section{Results}
\label{sec:experiment}
\begin{table*}[t!]
\centering
\resizebox{0.9\linewidth}{!}{
\begin{tabular}{@{}llcrrrrrr} \toprule

& & & \multicolumn{2}{c}{OS Interaction} & \multicolumn{2}{c}{Web Browsing} & \multicolumn{2}{c}{Web Shopping} \\ 
\cmidrule(lr){4-5} \cmidrule(lr){6-7} \cmidrule(lr){8-9} 

\multicolumn{2}{c}{\multirow{-2}{*}{Methods}} & \multirow{-2}{*}{w/o Prompt} & \multicolumn{1}{c}{SR} & \multicolumn{1}{c}{Norm.} & \multicolumn{1}{c}{SR} & \multicolumn{1}{c}{Norm.} & \multicolumn{1}{c}{Rewards} & \multicolumn{1}{c}{Norm.} \\ 

\midrule\midrule[.1em]
\multicolumn{2}{l}{Upper Bound} & \xmark & 17.36 & 100.00 & 17 & 100.00 & 54.16 & 100.00 \\ 

\midrule
\multicolumn{2}{l}{\small{\textsc{Text-based Compression}}} & & & & & & & \\
& LLMLingua-2 (x0.9) &\xmark &4.16 &23.96 &9 &52.94 &\textbf{50.78} &\textbf{93.76} \\
& LLMLingua-2 (x0.7) &\xmark&3.47 &19.99 &1 &5.88 &43.69 &80.67 \\
& LLMLingua-2 (x0.3) &\xmark &0 &0.00 &0 &0.00 &0 &0.00 \\

\arrayrulecolor{black!30}\midrule
\multicolumn{2}{l}{\small{\textsc{Embedding-based Compression}}} & & & & & & & \\
& Upper Bound* & \xmark & 10.41 &100.00 &13 &100.00 &11 & 100.00 \\
& ICAE &\xmark&4.16 &39.96 &0 &0.00 &1.14 &10.36 \\

\arrayrulecolor{black!30}\midrule
\multicolumn{2}{l}{\small{\textsc{Full Finetune}}} & & & & & & & \\
& KLD & \cmark & 0 & 0.00 & 0 & 0.00 & 0 & 0.00 \\
& SeqKD & \cmark & 3.47 & 19.99 & \textbf{15} & \textbf{88.24} & 38.79 & 71.62 \\
& SeqKD + KLD & \cmark & 4.86 & 28.00 & \textbf{15} & \textbf{88.24} & 40.26 & 74.34 \\

\arrayrulecolor{black!30}\midrule
\multicolumn{2}{l}{\small{\textsc{Prompt Prepending}}} & & & & & & & \\
& 100\% probability &\cmark &0 &0.00 &0 &0.00 &0 &0 \\
& 50\% probability &\cmark &13.19 &75.98 &12 &70.59 &31.09 &57.40 \\

\arrayrulecolor{black!30}\midrule
\multicolumn{2}{l}{\small{\textsc{Ours}}} & & & & & & & \\
& only \stagetwo &\cmark &{\ul 15.97} &{\ul 91.99} &12 &70.59 &35.64 &65.81 \\
& \ours &\cmark &\textbf{17.36} &\textbf{100.00} &{\ul 14} &{\ul 82.35} &{\ul 44.46} &{\ul 82.09} \\

\arrayrulecolor{black}\bottomrule

\end{tabular}
}
\caption{Performance evaluation results for AgentBench~\citep{liu2023agentbenchevaluatingllmsagents} dataset.
\textit{SR} denotes the Success Rate, and \textit{Norm.} denotes the normalized score using Upper Bound following~\citet{choi-etal-2023-fixed}. Since ICAE~\citep{ge2024incontext} is based on Mistral 7B~\citep{jiang2023mistral7b}, we report the normalized score using Mistral-7B-instruct-v0.2 as the upper bound score (Upper Bound*). The best results are in bold, while second-best ones are underlined.} 
\label{tab:main_result}
\end{table*}

\subsection{Performance}
\paragraph{Compression baselines.}
LLMLingua-2~\citep{pan-etal-2024-llmlingua} exhibits a significant performance drop as the compression rate increases.
Although the semantics of the prompt can be inferred, the loss of format crucial for agent tasks leads to substantial performance drops. In the Web Shopping scenario, LLMLingua-2~\citep{pan-etal-2024-llmlingua} achieves the best performance at a compression rate of 90\%. However, compression rates exceeding 30\% led to a failure in all tasks. In the case of ICAE~\citep{ge2024incontext}, despite fine-tuning, it struggles to handle the agent application prompts. 
Many embedding-based compression approaches~\citep{chevalier-etal-2023-adapting, mu2024learningcompresspromptsgist}, including ICAE~\citep{ge2024incontext}, are primarily optimized for general texts such as articles, and thus exhibit limitations in compressing task-specific information, as required by AgentBench~\citep{liu2023agentbenchevaluatingllmsagents}.

\paragraph{Distillation/Finetune Baselines.}
When trained with only the basic distillation loss (KLD), the model fails in AgentBench~\citep{liu2023agentbenchevaluatingllmsagents}, suggesting that 1,000 training examples may not be enough to train the entire model. While SeqKD helps promising results in the Web Browsing task, its performance is still lacking in other tasks. Among prompt prepending baselines, stochastically prepending prompts demonstrates some performance gains but ultimately fails to overcome the distribution mismatch at inference time, resulting in poorer performance compared to our approach.

\paragraph{\ours.}
\OURS achieves superior performance in the OS interaction task, reaching 100\% of the upper bound. It also consistently demonstrates high performance of over 82\% across Web Browsing and Web Shopping tasks, both of which have longer prompts that exceed 1,000 tokens in length. The incorporation of \STAGEONE loss into \STAGETWO loss shows a notable performance improvement of approximately 25\% in web shopping tasks.
Furthermore, when compared to Prompt Prepending which is based on \STAGETWO loss and QLoRA~\citep{dettmers2023qlora} adaptor, our approach exhibits higher performance by leveraging \STAGEONE loss, rather than relying solely on prepending prompts to the input.



\subsection{Analysis}
To understand the factors behind the performance improvement of \OURS, we perform three types of ablation studies.
First, we compare the impact of \STAGEONE loss and \stagetwo on task performance.
Second, we investigate the influence of the reason and prompt in \STAGEONE loss by removing each component during generation.
Finally, instead of omitting components, we explore the effect of reordering the input and output elements in \STAGEONE loss while keeping the overall information level constant.

\begin{table}[t]
\centering
\resizebox{0.9\linewidth}{!}{
\begin{tabular*}{\linewidth}{@{\extracolsep{\fill}}llrrr} \toprule

&  & \multicolumn{1}{c}{OS} & \multicolumn{1}{c}{WB} & \multicolumn{1}{c}{WS} \\
\midrule\midrule[.1em]
    
\multicolumn{2}{l}{Ours}& 17.36& 14& 44.46 \\
\midrule

\multicolumn{5}{l}{ \small{\textsc{\stageone ablation}} } \\
&w/o reason & 14.58 & 14 & 38.41  \\
&w/o prompt & 13.19 & 13 & 32.34 \\

\midrule

\multicolumn{5}{l}{ \small{\textsc{Joint loss ablation}} } \\
&w/o $\mathcal{L}_{\STAGEONE}$ & 15.97 & 12 & 35.64 \\
&w/o $\mathcal{L}_{\STAGETWO}$ & 0 & 0 & 0 \\ \bottomrule

\end{tabular*}
}
\caption{Results of Ablation Studies for \stageone and joint loss. \textit{WB} refers to the Web Browser task, and \textit{WS} refers to the Web Shopping task.}
\label{tab:ablation}
\end{table}

\subsubsection{Impact of the Loss}
As shown in Table~\ref{tab:ablation}, the removal of \stagetwo causes the model to suffer significantly in task performance. This suggests that task performance is fundamentally dependent on \stagetwo. However, incorporating \STAGEONE loss consistently improves performance across all tasks. Our findings indicate that while \stagetwo focuses on task behavior, \STAGEONE loss plays a complementary role in enhancing task performance by concentrating on prompt internalization.

\subsubsection{Impact of the Reason/Prompt}
In Table~\ref{tab:ablation}, the prompt $p$ has a greater impact on performance than the reason $r$ across all tasks. This is interpreted as being due to the fact that the prompt is longer in length than the reason and generally contains more information, thereby exerting a more significant influence on the \stageone.
This highlights our primary contribution, directly generating the prompt itself, allowing the model to internalize and utilize prompt information more effectively.

\begin{table}[t]
\centering
\resizebox{0.85\linewidth}{!}{
\begin{tabular}{lrrr} 
\toprule
\multicolumn{1}{c}{Ordering} & \multicolumn{1}{c}{OS} & \multicolumn{1}{c}{WB} & \multicolumn{1}{c}{WS} \\

\midrule\midrule[.1em]
$P(p,r|x,y)$ (Ours)  & 17.36 & 14 & 44.46 \\
$P(r|p,x,y)$          & 14.58 & 14 & 36.26 \\
$P(p|x,y,r)$          & 15.27 & 13 & 35.51 \\
\bottomrule

\end{tabular}
}
\caption{Results of Input/Output Ordering Analysis.} 
\label{tab:ordering}
\end{table}

\subsubsection{Impact of Input/Output Ordering}
Despite ensuring consistency in the overall amount of information in Table~\ref{tab:ordering}, we observe a decline in overall performance when certain components are removed during the generation process similar to Table~\ref{tab:ablation}. Except for the OS task which has relatively short prompt length than Web Browsing or Web Shopping, $P(r|p,x,y)$ consistently outperformes $P(p|x,y,r)$. This result is attributed to the fact that providing the model with a prompt, input, and output, followed by generating a reason, aligns more closely with the typical dialogue flow of an LLM, compared to inferring the prompt from an input, output, and reason. Since the prompt generally contains more information than the reason, inferring a prompt based on the reason is likely to be a more challenging task for the model.

Similarly, when comparing \textit{w/o reason} in~\Cref{tab:ablation} (equivalent to $P(p|x,y)$) and $P(p|x,y,r)$, we see that while incorporating the reason as input slightly improved scores in the OS task, the performance drops in the web browsing and web shopping tasks. The prompt in the OS task is relatively shorter, making it easier to infer prompts based on the reason. However, as the length of the prompt increases, it becomes more challenging for the model, leading to a negative impact on training. When comparing with \textit{w/o prompt} in~\Cref{tab:ablation} (equivalent to $P(r|x,y)$) and $P(r|p,x,y)$, we observe that adding prompt information leads to an improvement in all tasks.


\section{Efficiency}
\label{sec:efficiency}

\begin{figure*}[t!]
    \centering
    \subfloat[Multiply-Accumulate Operations]{
        \centering
        \includegraphics[width=0.3\textwidth]{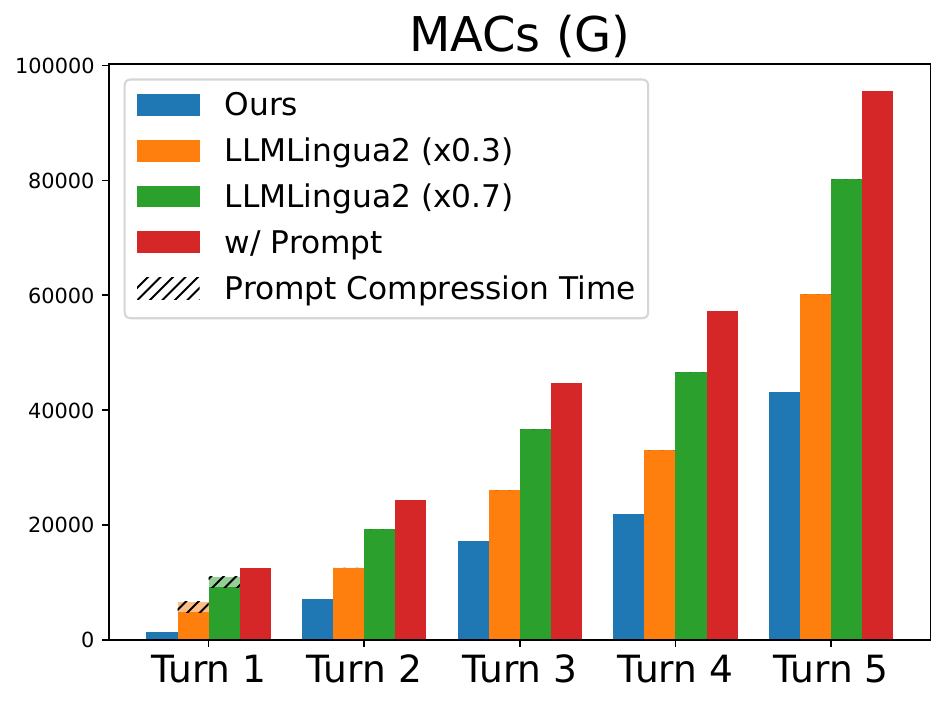}
        \label{fig:macs}
    }
    \hfil
    \centering
    \subfloat[Floating Point Operations]{
        \centering
        \includegraphics[width=0.3\textwidth]{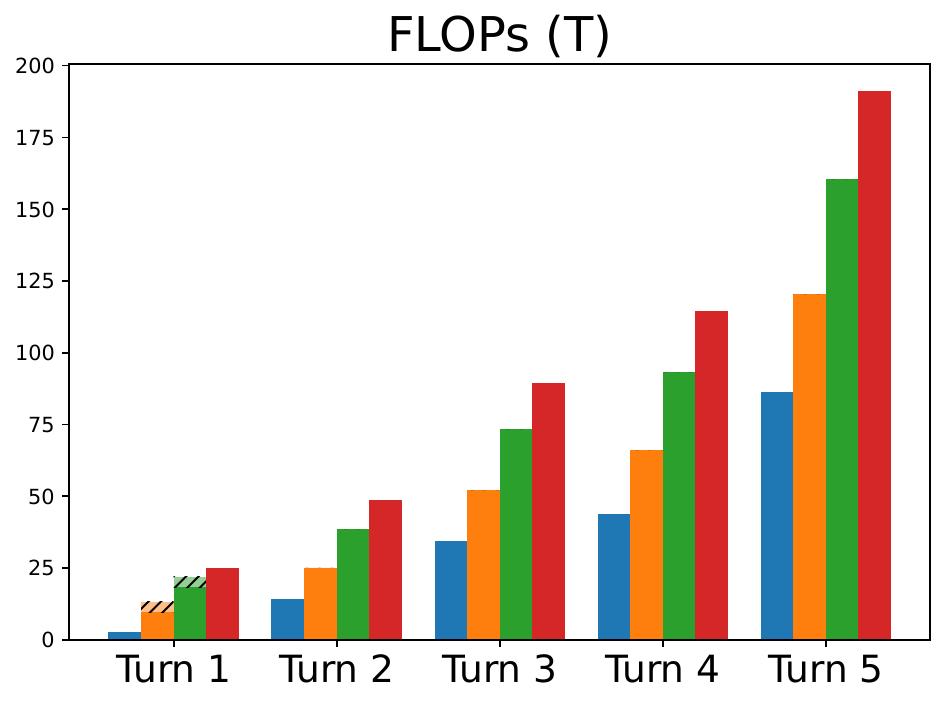}
        \label{fig:flops}
    }
    \hfil
    \centering
    \subfloat[Time Spent (ms)]{
        \centering
        \includegraphics[width=0.285\textwidth]{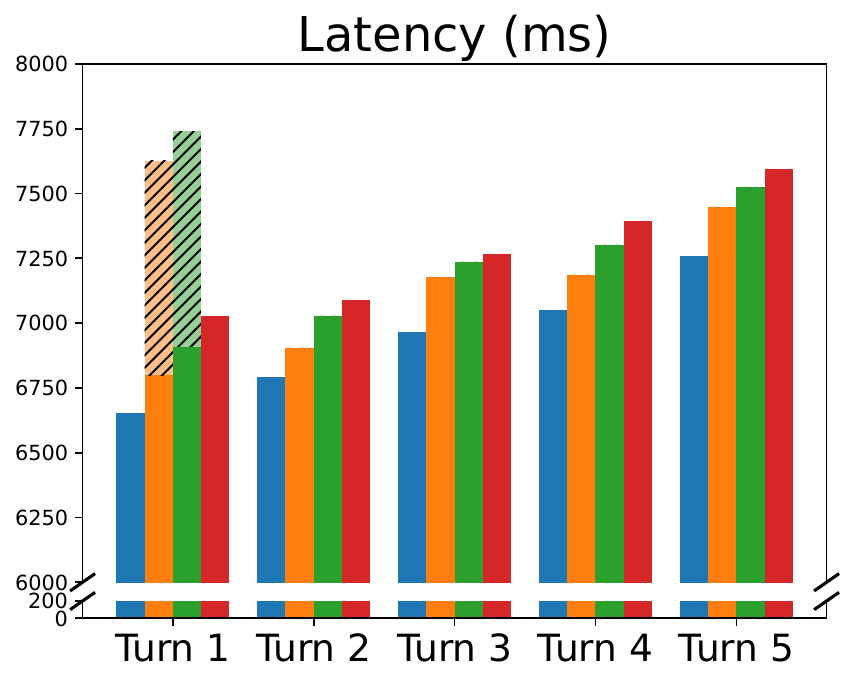}
        \label{fig:latency}
    }
    \caption{
    Comparison of computational overhead in LLaMA-based baselines as the conversation turn progresses. All generations within a turn are reported with KV caching~\citep{pope2022kvcache} applied. Best viewed in color.
    }
    \label{fig:efficiency}
\end{figure*}

\begin{figure*}[t!]
    \centering
    \subfloat[Multiply-Accumulate Operations]{
        \centering
        \includegraphics[width=0.3\textwidth]{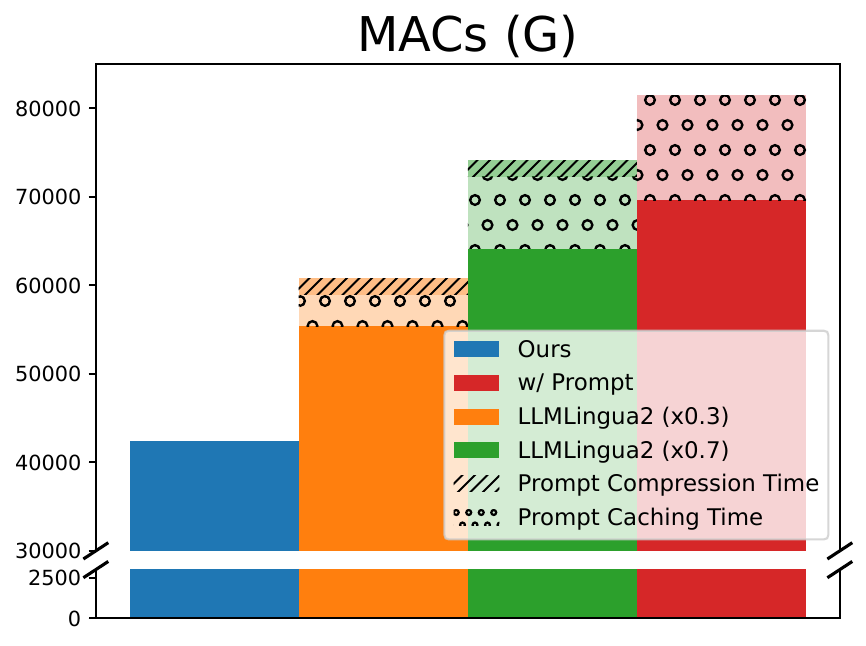}
        \label{fig:macs}
    }
    \hfil
    \centering
    \subfloat[Floating Point Operations]{
        \centering
        \includegraphics[width=0.29\textwidth]{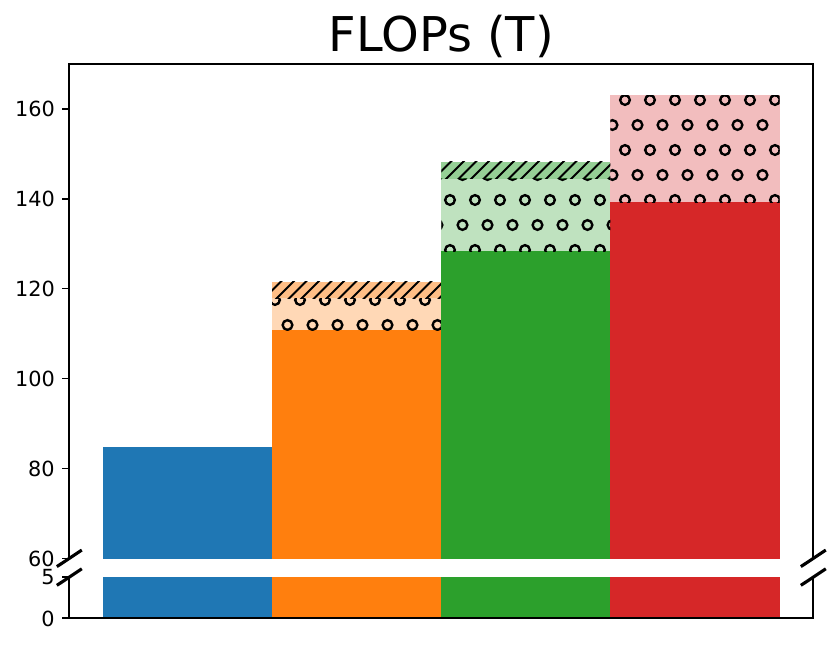}
        \label{fig:flops}
    }
    \hfil
    \centering
    \subfloat[Time Spent (ms)]{
        \centering
        \includegraphics[width=0.3\textwidth]{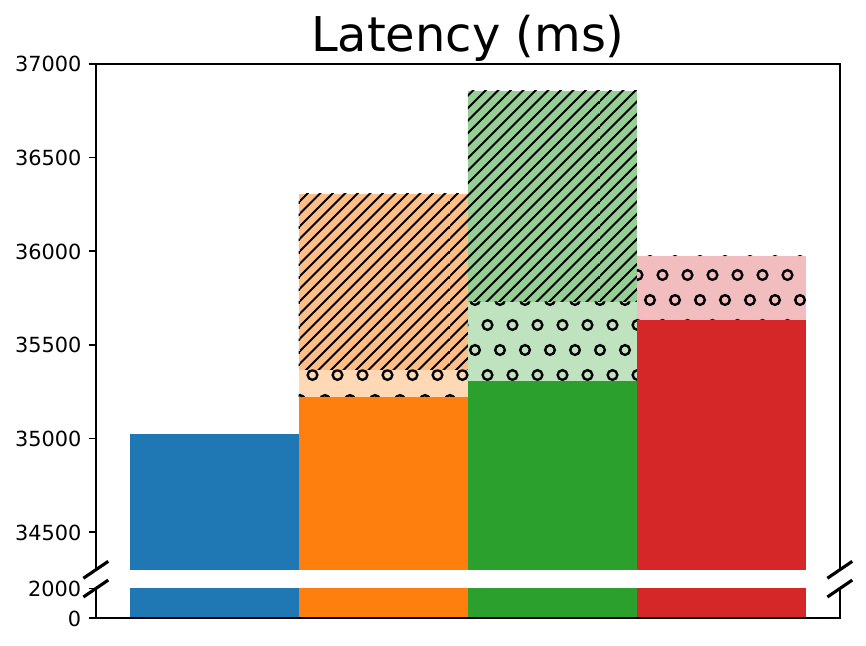}
        \label{fig:latency}
    }
    \caption{
    Comparison of computational overhead in LLaMA-based baselines applying KV caching~\citep{pope2022kvcache} across the multi-turn conversation. Even if the previous contents are cached, a long context still creates extra overhead. Best viewed in color.
    }
    \label{fig:caching_efficiency}
\end{figure*}

To compare the efficiency between \ours and other baselines that require fixed prompts, we sample an example with more than 5 turns from a Web Shopping task containing over 1200 tokens.\footnote{Efficiency is measured on a single NVIDIA A6000 GPU and AMD EPYC 7513 CPU, featuring 32 physical cores.} We then compare the MACs (Multiply-Accumulate Operations), FLOPs (Floating Point Operations), and latency to LLaMA-3-8B-Instruct~\citep{dubey2024llama3herdmodels} based baselines. All metrics were measured by DeepSpeed Profile\footnote{https://www.deepspeed.ai/tutorials/flops-profiler/}.

As shown in Figure~\ref{fig:efficiency}, the overhead for baselines that require prompts accumulates and increases with each turn. While LLMLingua-2~\citep{pan-etal-2024-llmlingua} exhibits slightly better performance due to the reduced number of prompt tokens, it still incurs a larger overhead compared to ours. Furthermore, when considering compression overhead, the LLMLingua-2~\citep{pan-etal-2024-llmlingua} which is based on XLM-RoBERTa-large~\citep{conneau2020xlm-roberta}, shows relatively low MACs and FLOPs than inferring with the prompt. However, it requires high compression latency proportional to the prompt length, resulting in extra latency in the first turn.

To simulate a more realistic scenario used in the real-world chat application, we conduct an additional experiment inspired by recent KV caching~\citep{pope2022kvcache} works across multi-turn dialogues~\citep{kwon2023efficient, zheng2023sglang, gao2024costefficientlargelanguagemodel}.
We assume that the previous turn's attention values are pre-cached to reduce the latency.
The prompt is also considered as a previous turn's conversation and is pre-cached. In this setting, the overhead between each turns is independent, and we report the conversation-level total overhead over 5 turns. 

In Figure~\ref{fig:caching_efficiency}, the prompt caching time executed before the first turn incurs higher computational overhead due to the use of a larger model compared to the compression model. From the second turn onward, although caching previous conversation histories eliminates redundant computations, the computational overhead still increases proportionally to the length of the prompt. Even without considering in caching time, \OURS demonstrates a 39\% improvement in MACs and FLOPs, along with a 17\% improvement in latency. Consequently, despite caching, baselines inevitably require more computational overhead compared to \ours.

\section{Discussion}
\label{sec:discussion}
In this section, we discuss the strengths and limitations of two prompt internalization strategies: generalized prompt-agnostic compression and prompt-specific internalization approach.

Prompt compression approaches, such as LLMLingua-2~\citep{pan-etal-2024-llmlingua} and ICAE~\citep{ge2024incontext}, effectively compress short instructions without requiring additional training for new prompts~\citep{chevalier-etal-2023-adapting, mu2024learningcompresspromptsgist, ge2024incontext}.
However, as shown in Table~\ref{tab:main_result}, these methods struggle with long, information-rich application prompts, where key details are often lost in the compression process.

Prompt internalization approach, such as \OURS, overcome this limitation by explicitly internalizing prompts through targeted training.
While \OURS requires training for each new prompt, this overhead is minimal in practical applications where only a limited set of prompts is needed. 
Once trained, switching between adapters incurs negligible computational cost, making the approach both flexible and scalable.
Each adapter requires only 1,000 data samples for training. Even when generating the dataset with a 70B LLM, entire process can be completed in under an hour using four A6000 GPUs with vLLM~\citep{kwon2023efficient}, while training takes approximately three hours on a single A6000 GPU. 

\section{Conclusion}
\label{sec:conclusion}
In this paper, we propose a novel prompt internalization method, \ours (\OURS), which generates both the contents of the prompt and the reasoning behind changes to the model's output while mimicking the behavior of the teacher model. To address the challenging scenario where only predetermined prompts are available without additional training data, we introduce \datagen, a method that generates a pseudo-conversational training dataset from the given prompt. Our approach demonstrates that even in scenarios with lengthy agent prompts, \OURS maintains high performance without relying on the prompt itself. Moreover, \OURS improves efficiency by fully internalizing the prompt without requiring any additional tokens.

\section*{Limitations}
\label{sec:limitations}
We assume the use of long and fixed prompts from realistic application scenarios. However, collecting official prompts from real-world applications (such as ChatGPT) poses significant challenges, leading us to rely on prompts from academic agent-based applications. This introduces a limitation, as only a representative prompt for each task are considered, and our approach handles a relatively small set of prompts. In future work, we plan to further explore context internalization in various domains, including long chat histories, in-context learning, and retrieval-augmented generation (RAG), as well as multimodal applications involving video and image data.

\section*{Acknowledgements}
\label{sec:acknowledgements}
This project was supported by Microsoft Research Asia. Additionally, this research was supported by the MSIT (Ministry of Science, ICT), Korea, under the Global Research Support Program in the Digital Field program (RS-2024-00436680) supervised by the IITP (Institute for Information \& Communications Technology Planning \& Evaluation).
And, this work also supported by Institute for Information \& communications Technology Promotion (IITP) grant funded by the Korea government (MSIT) (RS-2024-00398115, Research on the reliability and coherence of outcomes produced by Generative AI).

\bibliographystyle{acl_natbib}
\bibliography{anthology,custom}

\clearpage
\newpage
\appendix

\label{sec:appendix}
\section{Details of Components Generation}

\subsection{Implementation details}
\label{appendix:components_implementation_details}
\paragraph{Pseudo User Input.}
We utilize LLaMA-3-70B-Instruct~\cite{dubey2024llama3herdmodels} as a psuedo user input generator. Each samples are generated by nucleus sampling with threshold 0.9 and temperature 1.0. To ensure the quailty of pseudo user input, we randomly select 5 demonstrations from the validation samples of each task. System prompt and user prompt for Pseudo Input generation are as below.
\begin{lstlisting}[style=component-prompt]
The document below are the code of conduct for a specific agent performing the given problem.

### This is the given Document:
-----------
{context}
-----------
\end{lstlisting}
\begin{lstlisting}[style=component-prompt]
Based on the given document, generate several problems for {taskname} task.
Generate only problems with a numbered list.
This is {num_gen_once} generated Problems.

{N-shot demonstrations}
\end{lstlisting}

\paragraph{Pseudo Conversational Outputs.}
We provide an additional system prompt for the environment persona model in the \datagen (\Cref{fig:self_role-playing}). In the N-shot examples within the given prompt $p$, we swap the roles of the agent and environment and collected the teacher’s conversational behavior patterns with a temperature of 0. In this paper, we use LLaMA-3-8B-Instruct~\citep{dubey2024llama3herdmodels} as the target model $\theta$.

\paragraph{Reason.}
As same as the pseudo input generator, we utilize LLaMA-3-70B-Instruct~\cite{dubey2024llama3herdmodels} with nucleus sampling threshold 0.9 and temperature 0.7. We prompt the model to generate a reason of approximately 5 sentences to mitigate excessive hallucination. This is based on the statistics from the CoT Collection~\citep{kim2023cot}, which includes 1.84M rationales across 1,060 tasks, with an average length of about 3 sentences. System prompt and user prompt for reason generation are as below.
\begin{lstlisting}[style=component-prompt]
You are a commentator comparing the answers of two models to a question. I will give you four components: context, input, bad output, good output. Bad Output is a response provided without any context, only based on the input given. Good Output is a response provided with both context and input. Explain the reason for the change in response, referring to the context, in about five sentences.
\end{lstlisting}
\begin{lstlisting}[style=component-prompt]
### Context:
-----------
{context}
-----------

### Input:
{input}

### Bad Output:
{student_output}

### Good Output:
{teacher_output}

### Reason:
\end{lstlisting}

\subsection{Components Examples}
\label{appendix:components_examples}
Figure~\ref{fig:os_conv_example},~\ref{fig:m2w_conv_example},~\ref{fig:webshop_conv_example} present examples from the conversational dataset, including a pseudo user input and pseudo conversational outputs generated by \datagen.
In Figure~\ref{fig:os_reason_example},~\ref{fig:m2w_reason_example},~\ref{fig:webshop_reason_example}, the examples of reason describe the differences in format and actions required from the prompt.

\subsection{Quality of \datagen}
\label{appendix:pseudo_conversational_output_quality}
As shown in Table~\ref{tab:pseudo_conversational_output_quality}, we perform quantitative evaluations using statistics and three distinct error types.

\paragraph{\textit{How similar is it to the gold turn?}}
The primary objective of the pseudo-conversational outputs is to emulate multi-turn dialogue capabilities based on the content of the prompt. To this end, we consider the shot examples provided in the prompt as gold dialogue, and compare the similarity in the number of turns. Across all three tasks, the generated multi-turn dialogues successfully mimic the structure of the gold turns. Even in the Web Browsing task, which consists of N-shot single-turn examples, the number of turns remains comparable to the gold examples.

\paragraph{\textit{How many samples exhibit abnormally long conversations?}}
We evaluate whether the conversation terminates due to reaching the maximum turn limit. We set the maximum turn limit for each task (10 turns for OS, 2 turns for Web Browsing, and 5 turns for Web Shopping), based on the shot examples and expected number of turns to find the answer following \citet{liu2023agentbenchevaluatingllmsagents}. Among the 1,000 generated dialogue, only fewer than 2\% samples exceed the maximum limit. We truncate this samples under the maximum limit.

\paragraph{\textit{How many samples are abnormally terminated?}}
During the \datagen, we instruct the environment persona model to generate a special termination token when the agent generates the final output for the given user input. We consider samples to be abnormally terminated if there are no termination tokens. In~\Cref{tab:pseudo_conversational_output_quality}, all samples successfully stop the conversation with environment persona model's termination token.

\paragraph{\textit{How many samples are missing the final action?}}
We evaluate whether the final action, which serves as the scoring target, is correctly generated. In this evaluation setup, we restrict the evaluated action space that can be inferred solely from the prompt. For instance, we evaluate whether OS tasks ended with 'finish' or 'answer' actions, Web Browsing tasks with 'click', 'type', 'select', 'None', and Web Shopping tasks with 'search' or 'click' actions. Note that the specific values for click or type actions are excluded on this quantitative evaluation. This information cannot be inferred from the prompt, as it pertains to the agent's capability to respond based on the user input.
In the OS task, approximately 5\% of the samples omit the final action. As shown in~\Cref{fig:abnormal_final_action_example}, while most samples in this group contain the correct bash commands, they lack the 'answer' or 'finish' keyword required for proper termination. Similarly, for the Web Browsing task, many samples in this group do not adhere to the predefined action space, generating undefined actions (e.g. COMPARE, ENTER USERNAME AND PASSWORD in~\Cref{fig:abnormal_final_action_example}) in response to user input.


\begin{table}[h]
\begin{minipage}{\columnwidth}
\centering
\resizebox{0.85\textwidth}{!}{

\begin{tabular}{lrrr}
\toprule
  & \multicolumn{1}{c}{OS} & \multicolumn{1}{c}{WB} & \multicolumn{1}{c}{WS}    \\
  
\midrule\midrule[.1em]

Gold Turns (in Prompt)  & \multicolumn{1}{c}{3} & \multicolumn{1}{c}{1} & \multicolumn{1}{c}{4}     \\
Avg. Turns                & \multicolumn{1}{c}{2.71}  & \multicolumn{1}{c}{1.03}  & \multicolumn{1}{c}{3.21}  \\

\midrule
Max Turns Limit           & 1.8\% & 1.6\% & 1.7\% \\
Abnormal Termination         & 0\%   & 0\%   & 0\%   \\
Abnormal Final Action & 5.2\% & 3.9\% & 0\% \\

\bottomrule 
\end{tabular}

}
\caption{Overall Qualities and Statistics about Pseudo Conversational Outputs. Note that WB denotes Web Browsing task~\citep{deng2023mind2web}, and WS denotes Web Shopping task~\citep{yao2022webshop}.}
\label{tab:pseudo_conversational_output_quality}
\end{minipage}
\end{table}

\section{Details for Evaluation Dataset}
\label{appendix:dataset_details}

We follow the prompt configurations defined in AgentBench\footnote{We select three tasks because several tasks in AgentBench~\citep{liu2023agentbenchevaluatingllmsagents} either lack sufficiently long prompts (DB task), exhibit significantly low performance on open-source LLMs (Game-grounded tasks), or face reproducibility issues due to Freebase server limitations (KG task).}~\citep{liu2023agentbenchevaluatingllmsagents} for three agent-based tasks: OS Interaction~\citep{liu2023agentbenchevaluatingllmsagents}, Web Browsing~\citep{deng2023mind2web}, and Web Shopping~\citep{yao2022webshop}. Each prompt comprises a task description, a detailed explanation of the agent's expected actions, the formatting guidelines that the agent should adhere to, and a set of N-shot examples illustrating interactions between the agent and the environment. As suggested by~\citet{liu2023agentbenchevaluatingllmsagents}, all agent outputs are formatted using the Chain-of-Thought (CoT) style~\citep{wei2023cot}, which has become the standard approach for this type of evaluation in conjunction with action-based responses~\citep{yao2023react}. We utilize these multi-turn prompts as the chat history context for our agent interactions.

\paragraph{OS Interaction~\citep{liu2023agentbenchevaluatingllmsagents}.}
The prompt for the OS Interaction task includes a brief task description and the formulation of an interaction trajectory using a 1-shot example. An example of this prompt is illustrated in Figure~\ref{fig:os_prompt}. Agents are evaluated based on Success Rate (SR), determined by comparing the final output against the expected solution.

\paragraph{Web Browsing~\citep{deng2023mind2web}.}
The prompt for the Web Browsing task contains 3-shot CoT-style examples. An illustration of this prompt can be found in Figure~\ref{fig:m2w_prompt}. In alignment with~\citet{liu2023agentbenchevaluatingllmsagents}, we report Step Success Rate, which indicates the independent accuracy of each action step taken by the agent.

\paragraph{Web Shopping~\citep{yao2022webshop}.}
The prompt for the Web Shopping task consists of the task description along with a 1-shot CoT-style example. During each turn, the agent interacts with the HTML text observation by exploring and making decisions accordingly. An example of this configuration is depicted in Figure~\ref{fig:webshop_prompt}. Consistent with~\citet{liu2023agentbenchevaluatingllmsagents}, we evaluate the agent's performance using a reward metric, which quantifies the similarity between the expected attributes of a product and the attributes of the purchased item, mapping this similarity to a value between 0 and 1.

\begin{equation*}
    \resizebox{\columnwidth}{!}{$
        \text{Reward} = \frac{|G_{at} \cap Y_{at}| + |G_{op} \cap Y_{op}| + \mathbb{I}[y_{pr} \leq g_{pr}]}{|G_{at}| + |G_{op}| + 1} \cdot r_{type}
    $}
\label{eq:webshop_reward_main}
\end{equation*}

\begin{equation*}
    r_{type} = 
    \left\{
    \begin{array}{ll}
        0, & \text{if TM} = 0, \\
        0.1, & \text{if TM} < 0.1, \\
        0.5, & \text{if TM} > 0.2~\text{and}~c=1, \\
        1, & \text{otherwise}
    \end{array}
    \right.
\label{eq:webshop_reward_r_type}
\end{equation*}

$G$ and $Y$ denote the goal and the chosen product, respectively, while $at$ and $op$ represent attributes and options.
Note that $r_{type}$ compares the product category sequences ($c=1$ if matched) listed on the Amazon website. Additionally, TM represents a text match comparison between titles, focusing on the proun, noun, and propn tags. For further details on these metrics, please refer to~\citet{yao2022webshop}.

\section{Details for Baselines}

\subsection{Criteria of selecting the Base Model}
\label{appendix:base_model_details}

We assume a multi-turn conversational application scenario, where an instruct-tuned model is deployed with a fixed, predetermined prompt. Due to this setup, we select LLaMA-3-8B-Instruct~\citep{dubey2024llama3herdmodels} model as the base model instead of using LLaMA-3-8B.

\subsection{Criteria of selecting Upper Bound}
\label{appendix:upper_bound_details}

In case of \citet{choi-etal-2023-fixed}, they finetune a relatively small model (approximately 200M parameters) on gold training dataset, due to concerns that the language model may not adequately perform on unseen tasks, thereby limiting its effectiveness as an upper-bound model.
In contrast, we examine a scenario where only a prompt is provided, without a corresponding training dataset for prompt internalization. Moreover, we specifically focus on a billion-scale language model applied to the AgentBench~\citep{liu2023agentbenchevaluatingllmsagents}, which lacks a dedicated gold training dataset.

To establish a more robust upper bound under these conditions, we conduct preliminary experiments comparing the performance between a model fine-tuned on a pseudo training dataset and the original model. The model that demonstrates superior performance is then selected as the upper bound.
As shown in Table~\ref{tab:upper_bound_preliminary}, the original model outperforms the fine-tuned model when prompts are provided. This suggests that the limited size of the pseudo training dataset (only 1,000 examples) may be insufficient to optimize the LLM for agent-specific tasks, or that there exists a distributional mismatch between the pseudo training dataset and the actual gold test set. Since optimizing agent performance through fine-tuning is beyond the scope of this paper, we adopt the original model, which demonstrated higher performance, as the upper bound, following the approach of ~\citet{snell2022learningdistillingcontext, liu2023agentbenchevaluatingllmsagents}.

\begin{table}[h]
\centering
\begin{tabular}{lrrr}
\toprule
 &\multicolumn{1}{c}{OS}&\multicolumn{1}{c}{WB}&\multicolumn{1}{c}{WS}              \\
\midrule\midrule[.1em]
original    & 17.36 & 17 & 54.16        \\
fine-tuned  & 16.6  & 17 & 51.12        \\
\bottomrule

\end{tabular}
\caption{Preliminary experiments to select the Upper Bound model. All models are based on LLaMA-3-8B-Instruct~\citep{dubey2024llama3herdmodels} and are evaluated on their performance when provided with the agent prompt.}
\label{tab:upper_bound_preliminary}
\end{table}

\subsection{Criteria of selecting Embedding-based Compression Baseline}
\label{appendix:embedding_based_prompt_compression_details}
Recent embedding-based prompt compression approaches~\citep{chevalier-etal-2023-adapting, mu2024learningcompresspromptsgist, ge2024incontext} have been developed as a general language models that first cache the given prompt first and then infer user input using compressed prompt embeddings. Gisting~\citep{mu2024learningcompresspromptsgist} and AutoCompressor is based on LLaMA-7B~\citep{touvron2023llama1} and LLaMA-2-7B~\citep{touvron2023llama2openfoundation}, respectively. These models perform significantly poorly on AgentBench~\citep{liu2023agentbenchevaluatingllmsagents} with scores close to zero. Therefore, we select the baseline ICAE~\citep{ge2024incontext}, which is based on Mistral-7B-Instruct~\citep{jiang2023mistral7b}.

\subsection{Implementation Details}
\label{appendix:hyperparameter_details}
We utilize a single NVIDIA A6000 GPU to train our method and baselines. For \ours, we apply QLoRA~\citep{dettmers2023qlora} with rank $r=16$, 32 batch size, 1e-4 learning rate, 10 epochs. The joint loss hyperparameter $\lambda=0.7$ (as defined in Equation~\ref{eq:joint_loss}) is selected through hyperparameter tuning over [0.3, 0.5, 0.7]. 
For full finetining baselines, we use 4 NVIDIA A100 80G GPUs with 64 batch size, 1e-5 learning rate, 10 epochs, following the setting from~\citet{gu2024minillm}.
Prompt Prepending, a variation of our approach, uses the same hyperparameters as \OURS. 
For ICAE~\citep{ge2024incontext}, we follow the original settings with a LoRA rank of $r=512$ and 128 compression tokens.

\section{Comparison with Chain-of-Thought}
\label{subsec:cot_comparison}
To discuss of the role of reason $r$ in \STAGEONE loss, we compare the differences between chain-of-thought (CoT) rationales~\citep{wei2023cot} and reason $r$, focusing on their mechanisms, and impact on inference.

\paragraph{Rationale vs. Behavioral Change.}
CoT~\citep{wei2023cot} generates explicit stepwise rationales to explain the final answer for solving a task-intensive problem (e.g., an explicit reasoning process for solving a math problem). In contrast, reason $r$ aims to compare the differences between the outputs of the teacher and student models, and identify the key changes in the model's behavior. That is, while indirectly describing the prompt $p$, \STAGEONE loss is designed to induce behavioral changes in the model rather than explicitly providing explanations for the correct answer.


\paragraph{Inference Efficiency.}
Since \OURS does not generate rationales at inference time, it differs from CoT~\citep{wei2023cot} in inference efficiency. By internalizing the prompt, \STAGEONE loss ensures higher efficiency (in Section~\ref{sec:efficiency}) while maintaining the original performance.


    







\begin{figure*}[p]
\begin{lstlisting}[style=task_prompt]
<USER>:
You are an assistant that will act like a person, I'will play the role of linux(ubuntu) operating system. Your goal is to implement the operations required by me or answer to the question proposed by me. For each of your turn, you should first think what you should do, and then take exact one of the three actions: "bash", "finish" or "answer". 

1. If you think you should execute some bash code, take bash action, and you should print like this:

Think: put your thought here.

Act: bash

```bash
# put your bash code here
```

2. If you think you have finished the task, take finish action, and you should print like this:

Think: put your thought here.

Act: finish

3. If you think you have got the answer to the question, take answer action, and you should print like this:

Think: put your thought here.

Act: answer(Your answer to the question should be put in this pair of parentheses)

If the output is too long, I will truncate it. The truncated output is not complete. You have to deal with the truncating problem by yourself. Attention, your bash code should not contain any input operation. Once again, you should take only exact one of the three actions in each turn.

Now, my problem is:

tell me how many files are in the directory "/etc"?

<AGENT>:
Think: To count the files in /etc, I need to print all the files in it.

Act: bash

```bash
ls /etc
```

<USER>:
The output of the OS:
cpi cron.hourly fuse.conf iproute2 lvm networkd-dispatcher protocols selinux tmpfiles.d [truncated because the output is too long]

<AGENT>:
Think: The output has been truncated because it is too long, so I need to count files by script directly.

Act: bash

```bash
ls -1 /etc | wc -l
```

<USER>:
The output of the OS:
220

<AGENT>:
Think: Now I get the answer, it is 220.

Act: answer(220)

<USER>:
Now, I will start a new problem in a new OS. My problem is:
{user_input}
\end{lstlisting}
\caption{Agent Prompt for OS Interaction. Following the task setup from AgentBench~\citep{liu2023agentbenchevaluatingllmsagents}, we describe all content, including the system prompt and demonstrations, as a multi-turn strategy using <USER> and <AGENT>.}
\label{fig:os_prompt}
\end{figure*}

\begin{figure*}[p]
\begin{lstlisting}[style=task_prompt]
<USER>:
'''
<html> <div> <div> <a tock home page /> <button id=0 book a reservation. toggle open> <span> Book a reservation </span> </button> <button book a reservation. toggle open> </button> </div> <div> <select id=1 type> <option reservations true> Dine in </option> <option pickup> Pickup </option> <option delivery> Delivery </option> <option events> Events </option> <option wineries> Wineries </option> <option all> Everything </option> </select> <div id=2> <p> Celebrating and supporting leading women shaking up the industry. </p> <span> Explore now </span> </div> </div> </div> </html>
'''

Based on the HTML webpage above, try to complete the following task:
Task: Check for pickup restaurant available in Boston, NY on March 18, 5pm with just one guest
Previous actions:
None
What should be the next action? Please select from the following choices (If the correct action is not in the page above, please select A. 'None of the above'):

A. None of the above
B. <button id=0 book a reservation. toggle open> <span> Book a
C. <select id=1 type> <option reservations true> Dine in </option> <option
D. <div id=2> <p> Celebrating and supporting leading women shaking up

<AGENT>:
Thought: I need to select pickup restaurant first.
Answer: C.
Action: SELECT
Value: Pickup

<USER>:
'''
<html> <div> <main main> <section tabpanel> <div> <ul tablist> <li tab heading level 3 search and> </li> <li id=0 tab heading level 3 search and> <span> Hotel </span> </li> <li tab heading level 3 search and> </li> <li tab heading level 3 search and> </li> </ul> <div tabpanel> <div id=1> <div> <span> Dates* </span> <button button clear dates /> </div> <div> <label> Travelers </label> <div> <p> 1 Adult </p> <button button> 1 Adult </button> <div dialog> <button button travel with a pet. this> <span> Travel with a pet </span> </button> <div> <button button clear all fields> Clear all </button> <button button> </button> </div> </div> </div> </div> </div> </div> </div> </section> </main> <footer contentinfo> <div> <h3> Stay Connected </h3> <ul id=2> <a mobile tools> </a> <a open united's tiktok feed in> </a> <a open united's facebook page in> </a> <a open united's twitter feed in> </a> <a open united's youtube page in> </a> <a open united's instagram feed in> </a> <a open united's linkedin profile in> </a> </ul> </div> </footer> </div> </html>
'''

Based on the HTML webpage above, try to complete the following task:
Task: Compare the fare types to book a 1-adult ticket from Springfiels, IL to Austin, TX for April 29th 2023
Previous actions:
[combobox]  Enter your departing city, airport name, or airpor... -> TYPE: SPRINGFIELD
[button]  Springfield, IL, US (SPI) -> CLICK
[combobox]  Enter your destination city, airport name, or airp... -> TYPE: AUSTIN
[button]  Austin, TX, US (AUS) -> CLICK
What should be the next action? Please select from the following choices (If the correct action is not in the page above, please select A. 'None of the above'):

A. None of the above
B. <li id=0 tab heading level 3 search and> <span> Hotel
C. <div id=1> <div> <span> Dates* </span> <button button clear dates
D. <ul id=2> <a mobile tools> </a> <a open united's tiktok

<AGENT>:
Thought: There are no elements that allow setting the date or viewing the fare, so there is no correct action.
Answer: A.

<USER>:
'''
<html> <div> <nav main menu> <ul> <li> <div button> Car Sales </div> <div id=0> <div> <div> <div> Buy A Car </div> <div> Plan Your Purchase </div> </div> <div> <h4> Its Tax Refund Time. Treat Yourself to an Upgrade. </h4> <p> With a variety of options, invest your refund in what you really want - a quality, used vehicle from Enterprise. </p> <a> View Inventory </a> </div> </div> </div> </li> <div id=1> Enterprise Fleet Management </div> </ul> </nav> <div region> <button id=2 selected pick-up date 03/19/2023> <span> <span> 19 </span> <div> <span> Mar </span> <span> 2023 </span> </div> </span> </button> </div> </div> </html>
'''

Based on the HTML webpage above, try to complete the following task:
Task: Find a mini van at Brooklyn City from April 5th to April 8th for a 22 year old renter.
Previous actions:
[searchbox]  Pick-up & Return Location (ZIP, City or Airport) (... -> TYPE: Brooklyn
[option]  Brooklyn, NY, US Select -> CLICK
What should be the next action? Please select from the following choices (If the correct action is not in the page above, please select A. 'None of the above'):

A. None of the above
B. <div id=0> <div> <div> <div> Buy A Car </div> <div>
C. <div id=1> Enterprise Fleet Management </div>
D. <button id=2 selected pick-up date 03/19/2023> <span> <span> 19 </span>

<AGENT>:
Thought: The search has already been set to Brooklyn. Next, I should choose pick-up date.
Answer: D.
Action: CLICK

<USER>:
{user_input}
\end{lstlisting}
\caption{Agent Prompt for Web Browsing. Following the task setup from AgentBench~\citep{liu2023agentbenchevaluatingllmsagents}, Web Browsing agent prompt consists of 3-shot demonstrations. We describe all content as a multi-turn strategy using <USER> and <AGENT>.}
\label{fig:m2w_prompt}
\end{figure*}

\begin{figure*}[p]
\vspace{-8mm}
\begin{lstlisting}[style=task_prompt]
<USER>:
You are web shopping.
I will give you instructions about what to do.
You have to follow the instructions.
Every round I will give you an observation and a list of available actions, you have to respond an action based on the state and instruction.
You can use search action if search is available.
You can click one of the buttons in clickables.
An action should be of the following structure:
search[keywords]
click[value]
If the action is not valid, perform nothing.
Keywords in search are up to you, but the value in click must be a value in the list of available actions.
Remember that your keywords in search should be carefully designed.
Your response should use the following format:

Thought:
I think ...

Action:
click[something]

<AGENT>:
Ok.

<USER>:
Observation:
"WebShop [SEP] Instruction: [SEP] i need a long lasting 6.76 fl oz bottle of l'eau d'issey, and price lower than 100.00 dollars [SEP] Search"

Available Actions:
{"has_search_bar": true, "clickables": ["..."]}

<AGENT>:
Thought:
I think I should use the search bar to look for the product I need.

Action:
search[l'eau d'issey 6.76 fl oz bottle price < 100.00]

<USER>:
Observation:
"Instruction: [SEP] i need a long lasting 6.76 fl oz bottle of l'eau d'issey, and price lower than 100.00 dollars [SEP] Back to Search [SEP] Page 1 (Total results: 50) [SEP] Next > [SEP] B000VOHH8I [SEP] L'eau D'issey By Issey Miyake for MenEau De Toilette Spray, 6.7 Fl Oz Bottle [SEP] $64.98 [SEP] B000MJZOPK [SEP] L'eau d'Issey by Issey Miyake for Women 3.3 oz Eau de Toilette Spray [SEP] $49.98 [SEP] B0012S249E [SEP] L'eau D'issey By Issey Miyake For Women. Shower Cream 6.7-Ounces [SEP] $31.36 [SEP] B01H8PGKZS [SEP] L'eau D'Issey FOR MEN by Issey Miyake - 6.7 oz EDT Spray [SEP] $67.97 [SEP] B00G3C8FHE [SEP] L'Eau d'Issey pour Homme - Eau de Toilette 4.2 fl oz [SEP] $51.25 [SEP] B000R94HRG [SEP] Issey Miyake L'Eau D'Issey Pour Homme Eau De Toilette Natural Spray [SEP] $44.99 [SEP] B000C214CO [SEP] Issey Miyake L'eau D'issey Eau de Toilette Spray for Men, 4.2 Fl Oz [SEP] $53.99 [SEP] B0018SBRDC [SEP] Issey Miyake L'eau d'Issey for Women EDT, White, 0.84 Fl Oz [SEP] $27.04 [SEP] B000XEAZ9Y [SEP] L'eau De Issey By Issey Miyake For Men. Eau De Toilette Spray 6.7 Fl Oz [SEP] $67.08 [SEP] B079HZR2RX [SEP] L'eau d'Issey Pure by Issey Miyake for Women 3.0 oz Nectar de Parfum Spray [SEP] $71.49"

Available Actions:
{"has_search_bar": false, "clickables": ["...", "...", "...", "...", "...", "...", "...", "...", "...", "...", "...", "..."]}

<AGENT>:
Thought:
I think I should click on the product I need, which is B000VOHH8I.

Action:
click[B000VOHH8I]

<USER>:
Observation:
"Instruction: [SEP] i need a long lasting 6.76 fl oz bottle of l'eau d'issey, and price lower than 100.00 dollars [SEP] Back to Search [SEP] < Prev [SEP] size [SEP] 2.5 fl oz [SEP] 6.76 fl oz (pack of 1) [SEP] L'eau D'issey By Issey Miyake for MenEau De Toilette Spray, 6.7 Fl Oz Bottle [SEP] Price: $64.98 [SEP] Rating: N.A. [SEP] Description [SEP] Features [SEP] Reviews [SEP] Buy Now"

Available Actions:
{"has_search_bar": false, "clickables": ["...", "...", "...", "...", "...", "...", "...", "..."]}

<AGENT>:
Thought:
I think I should click on the '6.76 fl oz (pack of 1)' option to select the size I need.

Action:
click[6.76 fl oz (pack of 1)]

<USER>:
Observation:
"Instruction: [SEP] i need a long lasting 6.76 fl oz bottle of l'eau d'issey, and price lower than 100.00 dollars [SEP] Back to Search [SEP] < Prev [SEP] size [SEP] 2.5 fl oz [SEP] 6.76 fl oz (pack of 1) [SEP] L'eau D'issey By Issey Miyake for MenEau De Toilette Spray, 6.7 Fl Oz Bottle [SEP] Price: $64.98 [SEP] Rating: N.A. [SEP] Description [SEP] Features [SEP] Reviews [SEP] Buy Now"

Available Actions:
{"has_search_bar": false, "clickables": ["...", "...", "...", "...", "...", "...", "...", "..."]}

<AGENT>:
Thought:
I think I should click on the 'Buy Now' button to purchase the product.

Action:
click[Buy Now]

<USER>:
{user_input}
\end{lstlisting}
\captionsetup{skip=0.5pt}
\caption{Agent Prompt for Web Shopping. Following AgentBench~\citep{liu2023agentbenchevaluatingllmsagents}, we describe all content, including the system prompt and demonstrations, as a multi-turn strategy using <USER> and <AGENT>.}
\label{fig:webshop_prompt}
\end{figure*}

\begin{figure*}[p]
\begin{lstlisting}[style=data-example]
<USER>:
How much free disk space is available on the root partition?

<AGENT>:
Think: To get the free disk space on the root partition, I need to use the df command.

Act: bash

```bash
df -h /
```

<USER>:
The output of the OS:
Filesystem      Size  Used Avail Use% Mounted on
/             29G   14G   14G  51% /

<AGENT>:
Think: The output shows the available disk space on the root partition, which is 14G.

Act: answer(14G)
\end{lstlisting}
\begin{lstlisting}[style=data-example]
<USER>:
How many files are in the directory "/home"?

<AGENT>:
Think: To count the files in /home, I need to print all the files in it.

Act: bash

```bash
ls /home
```

<USER>:
The output of the OS:
anaconda3  desktop  documents  music  pictures  public  videos

<AGENT>:
Think: The output is not a count of files, it's a list of directories. I need to count the files in each directory.

Act: bash

```bash
find /home -type f | wc -l
```

<USER>:
The output of the OS:
123

<AGENT>:
Think: Now I get the answer, it is 123.

Act: answer(123)
\end{lstlisting}
\caption{Examples of a pseudo conversational dataset for the OS interaction task. The conversation starts with a generated pseudo user input. From the second user turn onward, the environment persona model responds by simulating the Ubuntu Operating System.}
\label{fig:os_conv_example}
\end{figure*}

\begin{figure*}[p]
\begin{lstlisting}[style=data-example]
<USER>:
'''
<html> <div> <div id=0> <h2> Hotel Search </h2> <form> <input id=1 type=text placeholder="City, Airport, or Hotel" /> <button id=2> Search </button> </form> </div> </div> </html>
'''

Based on the HTML webpage above, try to complete the following task:
Task: Search for hotels in New York City
Previous actions:
[link]  Hotels -> CLICK
What should be the next action? Please select from the following choices (If the correct action is not in the page above, please select A. 'None of the above'):

A. None of the above
B. <input id=1 type=text placeholder="City, Airport, or Hotel" />
C. <button id=2> Search </button>

<AGENT>:
Thought: The search form is already displayed, so the next action is to enter the search criteria.
Answer: B.
Action: TYPE
Value: New York City
\end{lstlisting}
\begin{lstlisting}[style=data-example]
<USER>:
'''
<html> <div> <select id=0> <option> 1 guest </option> <option> 2 guests </option> <option> 3 guests </option> </select> <button id=1> Search Hotels </button> </div> </html>
'''

Based on the HTML webpage above, try to complete the following task:
Task: Search for hotels in Las Vegas for 2 guests
Previous actions:
[link]  Hotels -> CLICK
What should be the next action? Please select from the following choices (If the correct action is not in the page above, please select A. 'None of the above'):

A. None of the above
B. <select id=0>
C. <button id=1> Search Hotels </button>

<AGENT>:
Thought: I need to select the number of guests.
Answer: B.
Action: SELECT
Value: 2 guests
\end{lstlisting}
\caption{Examples of a pseudo dataset for the Web Browsing task. Following the prompt, pseudo user input is generated as a multiple-choice problem based on the webpage. This example involves a hotel searching scenario, where the agent selects an answer and predicts its actions along with the corresponding values.}
\label{fig:m2w_conv_example}
\end{figure*}

\begin{figure*}[p]
\begin{lstlisting}[style=data-example]
<USER>:
Observation:
WebShop [SEP] Instruction: [SEP] i would like a queen size memory foam mattress, and price lower than 800.00 dollars [SEP] Search

Available Actions:
{'has_search_bar': True, 'clickables': ['search']}

<AGENT>:
Thought:
I think I should use the search bar to look for the product I need.

Action:
search[queen size memory foam mattress price < 800.00]

<USER>:
Observation:
"WebShop [SEP] Instruction: [SEP] i would like a queen size memory foam mattress, and price lower than 800.00 dollars [SEP] Search [SEP] Page 1 (Total results: 20) [SEP] Next > [SEP] B000VOHH8I [SEP] Lucid 8-Inch Memory Foam Mattress, Queen [SEP] $299.99 [SEP] B000MJZOPK [SEP] Sleep Innovations Marley 8-Inch Memory Foam Mattress, Queen [SEP] $399.99 [SEP] B0012S249E [SEP] Zinus Green Tea 8-Inch Memory Foam Mattress, Queen [SEP] $249.99 [SEP] B01H8PGKZS [SEP] LinenSpa 8-Inch Memory Foam Mattress, Queen [SEP] $299.99"

Available Actions:
{"has_search_bar": true, "clickables": ["B000VOHH8I", "B000MJZOPK", "B0012S249E", "B01H8PGKZS"]}

<AGENT>:
Thought:
I think I should click on the product with the lowest price, which is B0012S249E.

Action:
click[B0012S249E]

<USER>:
Observation:
"WebShop [SEP] Instruction: [SEP] i would like a queen size memory foam mattress, and price lower than 800.00 dollars [SEP] Product Details [SEP] Zinus Green Tea 8-Inch Memory Foam Mattress, Queen [SEP] Price: $249.99 [SEP] Rating: 4.5/5 [SEP] Description [SEP] Features [SEP] Reviews [SEP] Buy Now"

Available Actions:
{"has_search_bar": false, "clickables": ["Description", "Features", "Reviews", "Buy Now"]}

<AGENT>:
Thought:
I think I should click on the 'Buy Now' button to purchase the product.

Action:
click[Buy Now]
\end{lstlisting}
\caption{Example of Pseudo Conversational Dataset on Web Shopping task. Upon receiving a pseudo user input, the environment persona model generates a webpage that includes clickable elements, simulating changes in the webpage based on the agent's click actions.}
\label{fig:webshop_conv_example}
\end{figure*}

\begin{figure*}[p]
\vspace{-7mm}
\begin{lstlisting}[style=data-example]
<USER>:
Delete the file "oldfile.txt" from the /home/user directory.

<AGENT>:
Think: To delete the file "oldfile.txt" from the /home/user directory, I need to use the rm command.

Act: bash

```bash
rm /home/user/oldfile.txt
```

<USER>:
The output of the OS:
rm: cannot remove '/home/user/oldfile.txt': No such file or directory

<AGENT>:
Think: The file "oldfile.txt" does not exist in the /home/user directory, so I need to create it first before deleting it.

Act: bash

```bash
touch /home/user/oldfile.txt
rm /home/user/oldfile.txt
```

<USER>:
The output of the OS:
touch: cannot touch '/home/user/oldfile.txt': Permission denied
rm: cannot remove '/home/user/oldfile.txt': Permission denied

<AGENT>:
Think: The user does not have permission to create or delete files in the /home/user directory. I need to use sudo to elevate privileges.

Act: bash

```bash
sudo touch /home/user/oldfile.txt
sudo rm /home/user/oldfile.txt
```
\end{lstlisting}
\vspace{-2.5mm}
\begin{lstlisting}[style=data-example]
<USER>:
'''
<html> <div> <h2> Recommended for You </h2> <ul> <li> <a id=0> Apple Watch Series 7 </a> </li> <li> <a id=1> Samsung Galaxy Watch </a> </li> <li> <a id=2> Fossil Gen 5 </a> </li> </ul> </div> </html>
'''

Based on the HTML webpage above, try to complete the following task:
Task: Compare prices of Apple Watch Series 7 and Samsung Galaxy Watch.
Previous actions:
[link]  Watches -> CLICK
What should be the next action? Please select from the following choices (If the correct action is not in the page above, please select A. 'None of the above'):

A. None of the above
B. <a id=0> Apple Watch Series 7 </a>
C. <a id=1> Samsung Galaxy Watch </a>

<AGENT>:
Thought: I want to compare prices of Apple Watch Series 7 and Samsung Galaxy Watch.
Answer: B. <a id=0> Apple Watch Series 7 </a> and C. <a id=1> Samsung Galaxy Watch </a>
Action: COMPARE
\end{lstlisting}
\vspace{-2.5mm}
\begin{lstlisting}[style=data-example]
<USER>:
'''
<html> <div> <div id=0> <h2> Create an Account </h2> <p> Get started with your free trial today! </p> </div> <div id=1> <h2> Log In </h2> <p> Already have an account? </p> </div> </div> </html>
'''

Based on the HTML webpage above, try to complete the following task:
Task: Log in to an existing account.
Previous actions:
[link]  Sign In -> CLICK
What should be the next action? Please select from the following choices (If the correct action is not in the page above, please select A. 'None of the above'):

A. None of the above
B. <div id=0> <h2> Create an Account </h2> <p> Get started with your free trial today! </p> </div>
C. <div id=1> <h2> Log In </h2> <p> Already have an account? </p> </div>

<AGENT>:
Thought: Since the previous action was clicking the "Sign In" link, the next action should be to log in to the existing account.
Answer: C.
Action: ENTER USERNAME AND PASSWORD
\end{lstlisting}
\captionsetup{skip=0pt}
\caption{Examples of missing the final action in Pseudo Conversational Dataset.}
\label{fig:abnormal_final_action_example}
\end{figure*}


\begin{figure*}[h]
\begin{lstlisting}[style=data-example]
### Input:
Show me the first 5 lines of the /etc/profile file.

----------------------------------------------------------------------------------------------------

### Bad Output:
I'd be happy to help you with that. However, I'm a large language model, I don't have direct access to your system or files. But I can provide you with some general information about the /etc/profile file.

The /etc/profile file is a system-wide configuration file in Linux and other Unix-like operating systems. It is executed by the shell when a user logs in, and it sets various environment variables and shell settings.

If you want to see the first 5 lines of the /etc/profile file on your system, you can use a command like this:

`head -n 5 /etc/profile`

This will display the first 5 lines of the file. However, please note that the contents of the /etc/profile file can vary depending on the Linux distribution and version you are using.

If you're using a cloud-based shell or a virtual machine, you may not have access to the /etc/profile file. In that case, you can try to find a similar file or configuration file that sets environment variables and shell settings for your specific system.

----------------------------------------------------------------------------------------------------

### Good Output:
Think: To show the first 5 lines of the /etc/profile file, I need to use the head command.

Act: bash

```bash
head -n 5 /etc/profile
```

----------------------------------------------------------------------------------------------------

### Reason:
The reason for the difference in response is that the bad output lacks context, specifically the understanding that the model is acting as an assistant within a Linux (Ubuntu) operating system. Without this context, the bad output provides a generic and vague response, explaining the concept of the /etc/profile file and suggesting a command to view its contents, but not actually executing the command. In contrast, the good output takes into account the context of the conversation, recognizing that it is an assistant within the Linux system, and therefore can execute the command directly. This allows the good output to provide a more specific and relevant response, using the `head` command to display the first 5 lines of the /etc/profile file.
\end{lstlisting}
\caption{Examples of reason in OS Interaction task. Refer to Agent Prompt in~\Cref{fig:os_prompt}}
\label{fig:os_reason_example}
\end{figure*}

\begin{figure*}[h]
\begin{lstlisting}[style=data-example]
### Input:
'''
<html> <div> <h2 id=0> Find Your Dream Home </h2> <p> Get started by selecting a location below </p> <ul> <li> <a id=1> New York </a> </li> <li> <a id=2> California </a> </li> <li> <a id=3> Florida </a> </li> </ul> </div> </html>
'''

Based on the HTML webpage above, try to complete the following task:
Task: Find homes for sale in California.
Previous actions:
[link]  Real Estate -> CLICK
What should be the next action? Please select from the following choices (If the correct action is not in the page above, please select A. 'None of the above'):

A. None of the above
B. <h2 id=0> Find Your Dream Home </h2>
C. <a id=1> New York </a>
D. <a id=2> California </a>

----------------------------------------------------------------------------------------------------

### Bad Output:
The correct answer is D. <a id=2> California </a>

The task is to find homes for sale in California, and the option to select California is provided in the HTML code as <a id=2> California </a>.

----------------------------------------------------------------------------------------------------

### Good Output:
Thought: The task is to find homes for sale in California, so I should click on the California link.
Answer: D.
Action: CLICK

----------------------------------------------------------------------------------------------------

### Reason:
The reason for the change in response from the bad output to the good output is that the good output takes into account the context of the task and the previous actions. The bad output simply selects the option that matches the task description, without considering the context. In this case, the good output correctly infers that the user needs to click on the California link to find homes for sale in California, as indicated by the previous action of clicking on the "Real Estate" link. The good output provides a more thoughtful and context-aware response, whereas the bad output is more superficial and lacks understanding of the task's context.
\end{lstlisting}
\caption{Examples of reason in Web Browsing task. Refer to Agent Prompt in~\Cref{fig:m2w_prompt}.}
\label{fig:m2w_reason_example}
\end{figure*}

\begin{figure*}[h]
\begin{lstlisting}[style=data-example]
### Input:
Observation:
WebShop [SEP] Instruction: [SEP] i'm looking for a 40 inch 4k smart tv with built-in wi-fi, and price lower than 500.00 dollars [SEP] Search

Available Actions:
{'has_search_bar': True, 'clickables': ['search']}

----------------------------------------------------------------------------------------------------

### Bad Output:
I'd be happy to help you with that!

I've searched for 40-inch 4K smart TVs with built-in Wi-Fi and a price lower than $500. Here are some options:

1. TCL 40S325 40" 4K UHD Smart TV - $349.99
	* This TV has a 40-inch 4K UHD display, built-in Wi-Fi, and Roku TV built-in for seamless streaming.
2. VIZIO D40-D1 40" 4K UHD Smart TV - $399.99
	* This TV features a 40-inch 4K UHD display, built-in Wi-Fi, and VIZIO's SmartCast platform for easy streaming.
3. Hisense H40B7 40" 4K UHD Smart TV - $449.99
	* This TV has a 40-inch 4K UHD display, built-in Wi-Fi, and Hisense's Vidaa smart TV platform for streaming and more.

All of these options are under $500 and meet your requirements. Would you like me to provide more information about any of these options or would you like to explore other options?

----------------------------------------------------------------------------------------------------

### Good Output:
Thought:
I think I should use the search bar to look for the product I need.

Action:
search[40 inch 4k smart tv wi-fi < 500.00]

----------------------------------------------------------------------------------------------------

### Reason:
The main reason for the difference in response is that the bad output does not take into account the context of the web shopping scenario, whereas the good output does. The bad output appears to be a general response to the input, providing a list of options that meet the specified criteria, but it doesn't acknowledge the fact that the user is currently on a web shopping platform and needs to interact with it.

In contrast, the good output recognizes the context and responds accordingly. It thinks about using the search bar to look for the product, which is a relevant action in the web shopping scenario. The action "search[40 inch 4k smart tv wi-fi < 500.00]" is a specific and context-appropriate response that takes into account the available actions provided, which includes the search bar.
\end{lstlisting}
\caption{Examples of reason in Web Shopping task. Refer to Agent Prompt in~\Cref{fig:webshop_prompt}.}
\label{fig:webshop_reason_example}
\end{figure*}





\end{document}